# Economic span selection of bridge based on deep reinforcement learning

Leye Zhang[1], Xiangxiang Tian[1,*], Chengli Zhang[2], Hongjun Zhang[3]

(1. Jiangsu College of Finance & Accounting, Lianyungang, 222061, China;

2. Lianyungang Open University, Lianyungang, 222006, China;

3. Wanshi Antecedence Digital Intelligence Traffic Technology Co., Ltd, Nanjing, 210016, China)

**Abstract:** Deep Q-network algorithm is used to select economic span of bridge. Selection of bridge span has a significant impact on the total cost of bridge, and a reasonable selection of span can reduce engineering cost. Economic span of bridge is theoretically analyzed, and the theoretical solution formula of economic span is deduced. Construction process of bridge simulation environment is described in detail, including observation space, action space and reward function of the environment. Agent is constructed, convolutional neural network is used to approximate Q function, ε greedy policy is used for action selection, and experience replay is used for training. The test verifies that the agent can successfully learn optimal policy and realize economic span selection of bridge. This study provides a potential decision-making tool for bridge design.

## 0 Introduction

For a long bridge, laying aperture of bridge is an important task for bridge designers. The selection of bridge span has a huge impact on the total cost. When there are no rigid conditions such as navigable clearance, the bridge span should select the economic span, that is, the total cost of superstructure and substructure is the lowest. In terms of cost, the larger the span, the fewer the number of aperture of bridge, which can reduce the cost of substructure, but increase the cost of superstructure; On the contrary, the cost of superstructure will be reduced, but the cost of substructure will be increased. The terrain, geology, hydrology and other factors at the bridge site affect the value of economic span. For example, the value of economic span of sea-crossing bridges is usually greater than that of shoal of plain rivers. Laying aperture of bridge is a complex engineering problem, which needs to be analyzed and compared in terms of technology and economy, so as to obtain a more ideal design scheme[1-2].

Reinforcement learning (RL) originated from animal learning in psychology and optimization theory of optimal control in the 1950s and 1960s[3-4]. In 2016, AlphaGo defeated the human world champion of Go and officially brought reinforcement learning that has been developing silently for half a century into the public's vision. Based on deep reinforcement learning (DRL), Guozhong Cheng proposed an intelligent design method for high-rise shear wall structures[5]; Based on reinforcement learning, Jiachen He explored the active control of wind-induced vibration of long-span bridges[6]; Based on deep reinforcement learning, Quan Yuan proposed the intelligent railway location design approach[7]; Based on reinforcement learning, Ruifeng Luo proposed a generative design algorithm for truss structures[8]; Cheng M used deep reinforcement learning to put forward the decision-making framework for load rating planning of aging bridges[9]; Yang DY proposed a bridge management method considering asset and network risks based on deep reinforcement learning[10]. However, the application of reinforcement learning in economic span of bridge has not been reported.

In this paper, we build bridge simulation environment and use deep Q-network (DQN) algorithm of deep reinforcement learning to explore economic span selection of bridge (open source address of this article's https://github.com/zhangleye/BridgeSpan-DQN).

# 1 Theoretical analysis of economic span of bridge and introduction to deep reinforcement learning

## 1.1 Theoretical analysis of economic span of bridge

The idea of reference [2] is used for reference here, and appropriate simplification is made for the convenience of theoretical analysis (for example, the cost of two abutments is combined into the cost of one pier, and assume that the cost is a differentiable function of the span without considering the mutation caused by the change of structural form).

The total cost of bridge is the sum of the costs of superstructure and substructure.

$$W_{total}=W_{upper}+W_{under} \tag{1}$$

In the formula: $W_{total}$ is the total cost of bridge; $W_{upper}$ is the cost of superstructure; $W_{under}$ is the cost of substructure.

Superstructure consists of bridge deck system (pavement, anti-collision wall, etc.) and load-bearing structure. The cost of bridge deck system has nothing to do with span. It is assumed that the cost of load-bearing structure is a power function of span.

$$W_{upper}=(gS_1+C_1x^m S_2)L \tag{2}$$

In the formula: $gS_1$ is the cost of bridge deck system per unit length in the longitudinal direction, $g$ is the consumption of bridge deck system materials per unit length in the longitudinal direction, $S_1$ is the unit price of bridge deck system materials; $C_1x^m S_2$ is the cost of load-bearing structure per unit length in the longitudinal direction, $C_1x^m$ is the material consumption of load-bearing structure per unit length in the longitudinal direction, $C_1$ is the coefficient of power function, $x$ is the span (independent variable of power function), $m$ is the exponent of power function (not less than 1), $S_2$ is the unit price of load-bearing structure material; $L$ is the total length of all spans.

It is assumed that the cost of substructure is also a power function of span.

$$W_{under}=\frac{L}{x}C_2x^{1/n}S_3 \tag{3}$$

In the formula: $\frac{L}{x}$ is the number of piers; $C_2x^{1/n}S_3$ is the cost of a single pier, $C_2x^{1/n}$ is the material consumption of a single pier, $C_2$ is the coefficient of the power function, $\frac{1}{n}$ is the exponent of the power function (n>1), $S_3$ is the unit price of pier material.

Derivative of total bridge cost with respect to span:

$$\frac{d(W_{total})}{dx}=C_1mx^{m-1}S_2L+(\frac{1}{n}-1)C_2x^{1/n-2}S_3L \tag{4}$$

The function graph of total bridge cost is concave. When the derivative is 0, the total bridge cost is the lowest, that is:

$$C_1mx^{m-1}S_2L=(1-\frac{1}{n})C_2x^{1/n-2}S_3L \tag{5}$$

$$C_1x^{m+1}S_2=\frac{n-1}{mn}C_2x^{1/n}S_3 \tag{6}$$

In the formula:: $C_1x^{m+1}S_2$ is the cost of single span load-bearing structure.

That is, when the cost of single span load-bearing structure = $\frac{n-1}{mn}$ the cost of single pier, it is the economic span.

## 1.2 Introduction to deep reinforcement learning

Reinforcement learning is a branch of machine learning. It addresses what kind of action an agent should take in an environment in order to maximize the reward. The environment has built-in rules for status update and reward, but these rules are a black box for agents. Agent can only interact with the environment through Trail-and-error method, and find the optimal action policy through the feedback of the environment. For example, in a bridge simulation environment, agent first randomly selects load-bearing structural material and span, and then tries to adjust the material and span. Some adjustments will increase the cost, while others will reduce the cost. Through continuous attempts and summaries, it finally finds the material and span with the lowest bridge cost.

There are many algorithms that can help agent find the optimal action policy. Among them, Q-learning algorithm has a far-reaching impact and lays the foundation for many later algorithms. The specific steps of the Q-learning algorithm are as follows: first, create a table (Q value table, the number of rows is the number of states, the number of columns is the number of actions, and the cell value is the value of executing the action in this state). At the beginning, the cell value (Q value) of the table is a random value; Then the agent interacts with the environment, and updates each Q value in the table step by step according to the feedback of the environment. Through multiple iterations, all Q values are almost unchanged. With the correct Q-value table, the agent will know what action to perform under what state, so as to achieve the goal.

Deep reinforcement learning combines the perception ability of deep learning (DL) with the decision-making ability of reinforcement learning, and controls the action according to the input state image. This end-to-end learning method is closer to human thinking. DQN is a mainstream and widely used deep reinforcement learning algorithm. DQN essentially uses convolutional neural network to replace the Q value table. The specific steps are as follows: first, establish convolutional neural network (the input is the state image, and the output is the Q value of various actions executed in this state). At the beginning, the weight parameters of neural network are random value. Then the agent interacts with the environment, and updates the weight parameters of the neural network gradually according to the feedback of the environment. Through multiple iterations, all outputs of the neural network are almost unchanged. With the correct weight parameters of neural network, the correct Q value can be output for each state, and the agent will know what action to perform under what state, so as to achieve the goal.

## 2 Economic span selection of bridge based on deep reinforcement learning

To solve practical engineering problems by reinforcement learning, we should first establish the environment, then establish the agent, and finally train and test.

This task is based on Python 3.10 programming language, OpenAI Gym 0.26.2 reinforcement learning toolkit, TensorFlow 2.10 and Keras2.10 deep learning platform framework.

### 2.1 Self-built bridge simulation environment

Since almost no self-built environment is involved in reinforcement learning papers and books on the market, a detailed process is given here.

1. the environment is a two-dimensional grid with 3 rows and 80 columns, and the coordinate origin is located in the upper left corner. The row represents the material category of load-bearing structure, and the rows 0, 1 and 2 from top to bottom are concrete structure, steel-concrete composite structure and steel structure in turn. Columns represent spans, and columns 0, 1, 2,..., 77, 78 and 79 from left to right are 10, 20, 30,..., 780, 790 and 800 meters in sequence. Each cell is distinguished by black and gray alternately. The cell where the agent is located is represented by red, and the number of observation spaces is 3 × 80 = 240. See Figure 1 for details.

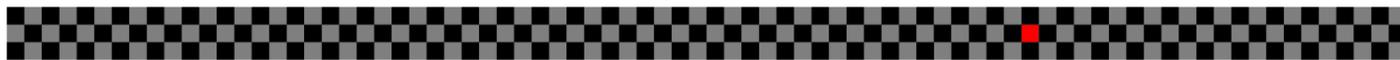

Fig.1 Schematic diagram of bridge simulation environmen

The number of action spaces in the environment is 5, which are 0 (standing still), 1 (one grid up), 2 (one grid down), 3 (one grid left), and 4 (one grid right).

The agent can do any action in any cell in the environment (if it runs out of the environment, it will turn back). In Figure 1, the agent is located in row 1 and column 58. If the "1" action is executed, the position after the action is row 0 and column 58. Agents constantly move in the environment and find the cell of economic span by maximizing reward.

See Figure 2 for the environmental attribute code.

```python
class BridgeSpanEnv(gym.Env):
    def __init__(self, Min_length=10,Max_length=800,step_length=10,max_steps=200):
        self.Min_length=Min_length
        self.Max_length=Max_length
        self.step_length=step_length
        self.max_steps=max_steps
        self.Material_Type=3
        self.state_pos_dict = np.arange(self.Min_length,self.Max_length+self.step_length,self.step_length)
        self.observation_space = gym.spaces.Discrete( self.Material_Type*len(self.state_pos_dict) )
        self.states=np.arange(self.observation_space.n)
        self.state=random.choice(self.states)
        self.action_space = gym.spaces.Discrete(5)
        self.actions = [NOOP, UP, DOWN, LEFT, RIGHT]
        self.action_pos_dict = {NOOP:[0,0],UP:[-1,0],DOWN:[1,0],LEFT:[0,-self.step_length],RIGHT:[0,self.step_length]}
        self.step_num = 0
        self.done = False
        self.truncated= False
        self.info = {}
        self.viewer = None
        self.img_shape = [16*self.Material_Type, 16*len(self.state_pos_dict), 3]
```

Fig.2 Code for environment properties

2. the step() method is the core of the environment. It receives the action of the agent as a parameter and returns the state, reward and other information.

It first calculates the row and column coordinate and state after the action, and then calculates the reward of the action.

Solving the optimal policy is the only task of reinforcement learning. The agent maximizes the reward by improving the policy, so the design of the reward function is the key. This task is to find the economic span of bridge, and the lowest total cost is the goal. Therefore, actions that can reduce the cost produce high reward, and actions that can increase the cost produce low reward, that is, the cost is negatively correlated with the reward. So the negative value of the total bridge cost is used as the reward.

Based on formulas (1), (2), (3) and engineering experience, it is assumed that the calculation formula for the cost of superstructure and substructure per unit area of bridge is(Unit: Yuan per square meter): ① concrete structure: the unit price of superstructure is $250+40x^{1.2}$, and the unit price of substructure is $50000x^{-0.5}$. ② Steel-concrete composite structure: the unit price of superstructure is $500+90x^{1.07}$, and the unit price of substructure is $45000x^{-0.5}$. ③ Steel structure: the unit price of superstructure is $2000+140x^{1.0}$, and the unit price of substructure is $40000x^{-0.5}$. (Note: Calculated manually according to formula (6), the economic span of concrete structure, steel-concrete composite structure and steel structure and the total cost of per unit area of the corresponding bridge are 39.6 m/11501 yuan per square meter, 32.3 m/12125 yuan per square meter and 27.3 m/13478 yuan per square meter respectively. Taking the lowest cost of the three, the economic span of the environment is 40m/concrete structure, corresponding to the cell in row 0 and column 3 of the environment, which is the destination of the agent. If the agent wanders in the environment and can always spend less steps to reach and stay at the destination, it means learning success.)

See Figure 3 for relevant codes.

```python
def step(self, action):
    number_of_rows, span=self.state_to_grid(self.state)
    next_number_of_rows = number_of_rows + self.action_pos_dict[action][0]
    next_span = span + self.action_pos_dict[action][1]
    if next_number_of_rows < 0 or next_number_of_rows > 2:
        next_number_of_rows =number_of_rows
    if next_span < self.Min_length or next_span > self.Max_length:
        next_span =span
    self.state=self.grid_to_state(next_number_of_rows, next_span)
    self.info = {"number_of_rows(0=concrete、1=composite、2=steel)": next_number_of_rows,"span(m)":next_span}
    if next_number_of_rows==0:
        reward= (40*next_span**1.2+250)+(next_span**(1/2-1)*50000)
    elif next_number_of_rows==1:
        reward= (90*next_span**1.07+500)+(next_span**(1/2-1)*45000)
    elif next_number_of_rows==2:
        reward= (140*next_span**1+2000)+(next_span**(1/2-1)*40000)
    self.step_num += 1
    if self.step_num >= self.max_steps:
        self.done = True
    return self.state, -reward, self.done, self.truncated, self.info
```

Fig.3 Code for step() method of Environment

3. The reset() method is used to reset the environment and randomly assign an initial position (cell) to the agent. The state_to_grid() and grid_to_state() methods are used to convert the state index to the row and column coordinate. See Figure 4 for relevant codes.

```python
def reset(self):
    self.state=random.choice(self.states)
    number_of_rows, span=self.state_to_grid(self.state)
    self.info = {"number_of_rows(0=concrete、1=composite、2=steel)": number_of_rows,"span(m)":span}
    self.step_num = 0
    self.done = False
    self.truncated= False
    return self.state, self.info

def state_to_grid(self,the_state):
    number_of_rows=the_state//len(self.state_pos_dict)
    span=self.state_pos_dict[the_state%len(self.state_pos_dict)]
    return number_of_rows, span

def grid_to_state(self,number_of_rows, span):
    the_state=np.where(self.state_pos_dict == span)[0][0]+number_of_rows*80
    return the_state
```

Fig.4 Code for reset(), state_to_grid(), grid_to_state() method of Environment

4. the render() and gridarray_to_image() methods are used to render state images. See the source code for details.

## 2.2 Construction and training of agent

The establishment and training of DQN agent is a routine operation. Reinforcement learning books on the market have detailed descriptions[3-4], so here is only a brief introduction. See the source code for relevant details.

1. Establish convolutional neural network

The input is the state image, and the output is the Q value of various actions executed in this state.

The convolutional base is stacked by two Conv2d layers, and the number of filters is 16 and 32 in turn. The sampling window size is 4 × 4, the stride size is 4, and the activation function is relu.

The classifier is stacked by two Dense layers, the number of neurons is 128 and 5, and the activation function is relu and linear.

Tab.1 Model summary of convolutional neural network

| Layer (type) | Output Shape | Param # |
| --- | --- | --- |
| conv2d (Conv2D) | (None, 12, 320, 16) | 784 |
| conv2d_1 (Conv2D) | (None, 3, 80, 32) | 8224 |
| flatten (Flatten) | (None, 7680) | 0 |
| dense (Dense) | (None, 128) | 983168 |
| dense_1 (Dense) | (None, 5) | 645 |
| Total params: 992,821 | | |
| Trainable params: 992,821 | | |
| Non-trainable params: 0 | | |

2. Use greedy policy to select action: select random action with probability $\varepsilon$, and select action with maximum Q value with probability $(1-\varepsilon)$.

3. Establish experience replay cache in queue mode to save the data of current state, action, return and next state.

4. Replay the data in the experience replay cache and train the convolutional neural network. Label is the total reward of environmental feedback, which is composed of immediate reward and next state value (bootstrap). The purpose of training is to make the predicted value of the model match the label. The loss function uses mean square error (MSE), and the optimizer uses "Adam". Finally, the convolutional neural network can output the correct Q value for each state. The training loss curve is shown in the following figure (the first 100 rounds):

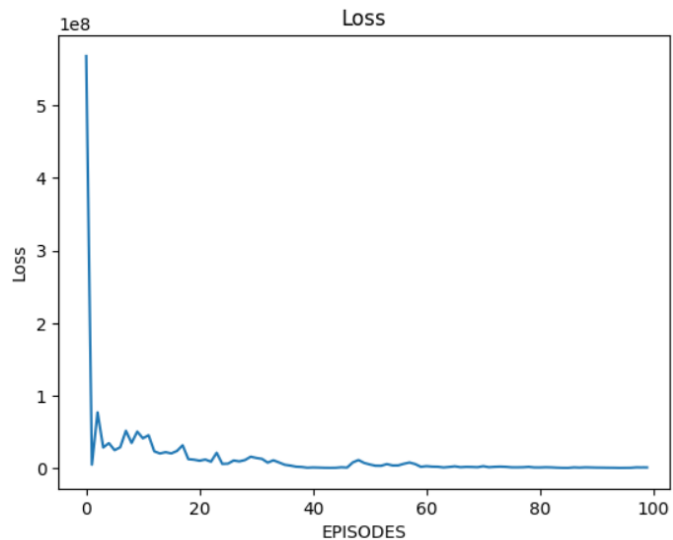

Fig.5 Training loss curve

## 2.3 Testing

Q value is the total reward of the action. A successfully trained agent can successfully achieve the goal by executing the action with the maximum Q value in each state. For this environment, no matter which cell the agent is initially assigned to, it should be able to quickly move to the cell in row 0 and column 3 (economic span). The test results are shown in the following figure (in the figure, the red cell is the initial position of the agent, the blue cell is the motion trajectory of the agent, and the green cell is the motion end point of the agent):

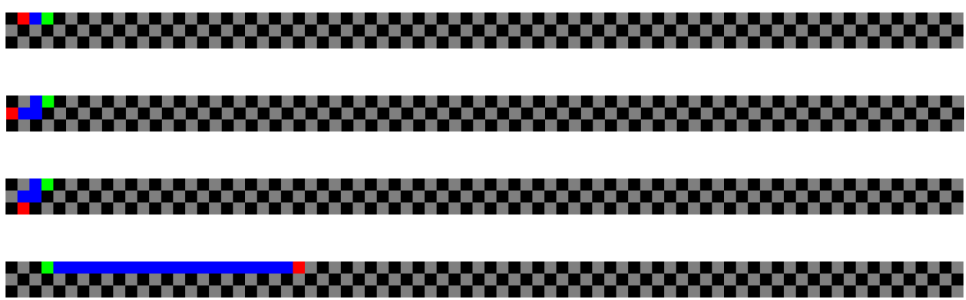

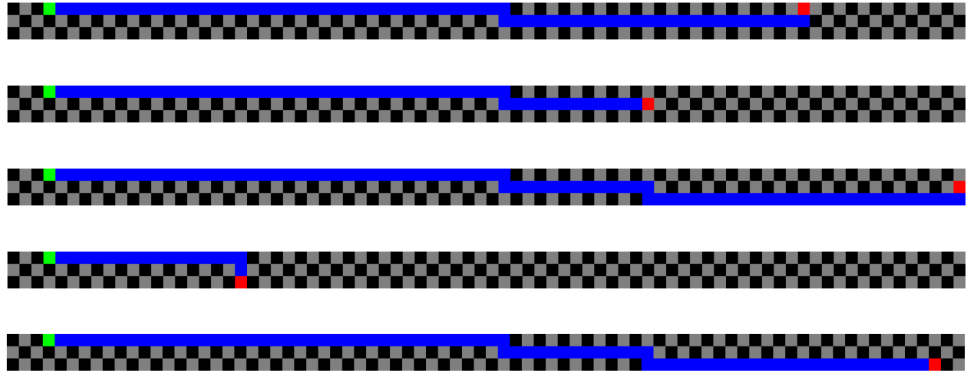

Fig.6 Policy testing of agent

The above figure shows that all end points of the agent are the cell in row 0 and column 3 (economic span). It can be seen that the trained agent can successfully select economic span of bridge.

## 3 Conclusion

In this paper, derivative method for finding the extreme value of a function is used to deduce the theoretical solution formula of economic span.

By constructing bridge simulation environment and applying deep Q-network algorithm, the application of deep reinforcement learning in economic span selection of bridge is successfully realized. The result shows that the agent can quickly and effectively identify bridge span with the lowest cost through learning, which provides a new perspective for the field of bridge design, and also provides a useful exploration for the application of reinforcement learning in engineering problems.

There are some limitations in this attempt, such as the cost function, the change of bridge structure is not considered in bridge simulation environment, and training time of agent is too long, which can be further optimized in the future.